%% file: main.tex
\documentclass[runningheads]{llncs}

\usepackage{eccv}

\usepackage{eccvabbrv}

\usepackage{graphicx}
\usepackage{booktabs}

\usepackage[accsupp]{axessibility}  

\usepackage{hyperref}

\usepackage{orcidlink}

\definecolor{cvprblue}{rgb}{0.21,0.49,0.74}

\usepackage{multirow}
\PassOptionsToPackage{nameinlink,capitalise,capitalize}{cleveref}
\usepackage{hyperref}
\hypersetup{breaklinks,colorlinks,allcolors=cvprblue}
\usepackage{cleveref}
\usepackage{xcolor}
\usepackage[utf8]{inputenc}
\definecolor{turquoise}{rgb}{0.0,0.5,0.5}

\definecolor{orange}{RGB}{254,180,43} 

\usepackage{placeins}
\usepackage{subcaption}
\usepackage{amsmath}
\usepackage{amssymb}
\usepackage{booktabs}
\usepackage{soul} 
\usepackage{amsfonts}
\usepackage{stmaryrd}
\usepackage{dsfont}
\usepackage{xfrac}
\usepackage{mathrsfs}
\usepackage{dsfont}
\usepackage{multirow}
\usepackage{tabularx}
\usepackage{tabulary}
\usepackage{tcolorbox}
\usepackage{colortbl}
\tcbuselibrary{most}
\usepackage{setspace}
\usepackage[bb=dsserif]{mathalpha}
\usepackage{makecell}
\usepackage{float}
\usepackage{amsmath}
\usepackage{ifthen}
\usepackage{pifont}
\usepackage[table]{xcolor}

\graphicspath{{../CVPR/}}
\usepackage{graphicx}
\usepackage{svg}
\usepackage{adjustbox}
\usepackage{longtable}
\usepackage{ulem} 

\makeatletter
\renewcommand\paragraph[1]{%
  \vspace{0.5em}%
  \noindent\textbf{\textit{#1}}
}
\makeatother

\begin{document}

\title{Who Made This? Fake Detection and Source Attribution with Diffusion Features} 

\titlerunning{Who Made This?}

\author{Simone Bonechi\inst{1} \and
Paolo Andreini\inst{1} \and
Barbara Toniella Corradini \inst{2}}

\authorrunning{S.~Bonechi et al.}

\institute{Department of Information Engineering and Mathematics, University of Siena, Siena, Italy 
\and
AIGO, Italian Institute of Technology, Genoa, Italy
}

\maketitle

\begin{center}
    \centering
    \captionsetup{type=figure}
    \includegraphics[width=.90\textwidth]{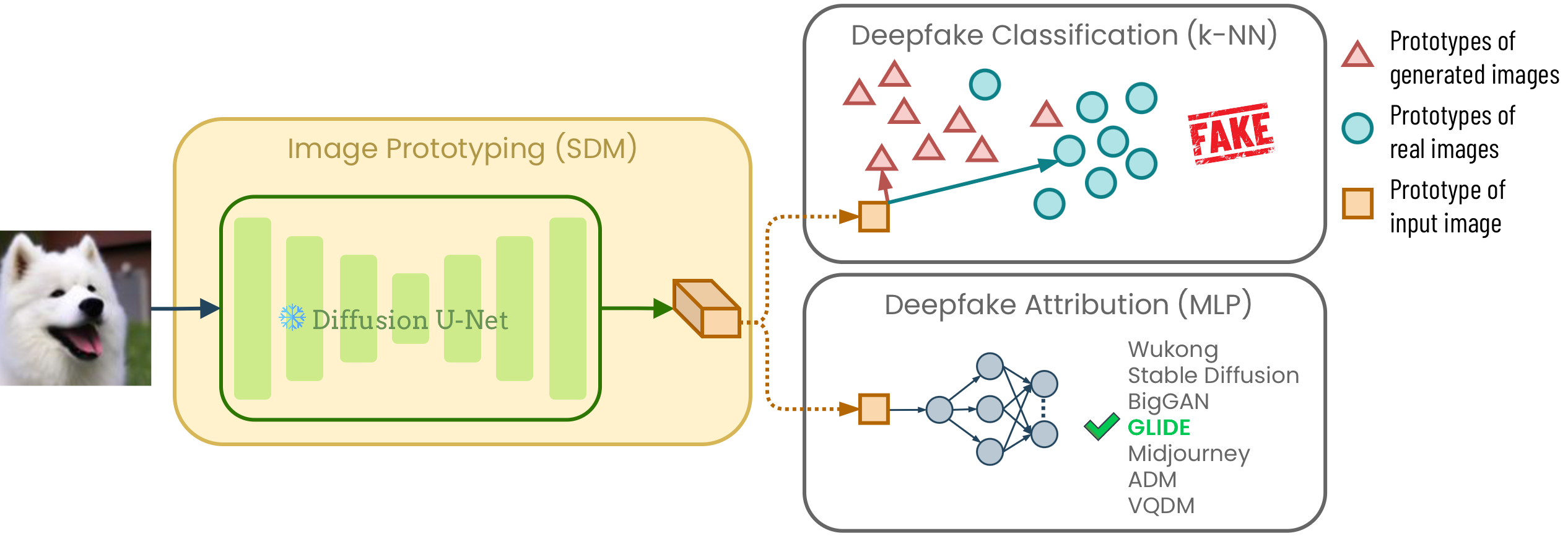}
    \caption{\textbf{Overview of our framework for fake image classification and source model attribution.} 
    Given an input image, our Image Prototyping module extracts image prototype from the U-Net of a Stable Diffusion Model. 
    This compact representation is then used for two downstream tasks: (i) fake image classification via a $k$-NN classifier to determine whether the image is real or fake, and (ii) source model attribution via an MLP to identify the generative model (e.g., GLIDE, Stable Diffusion, BigGAN) responsible for the fake.}
    \label{fig:main}
\end{center}%

\input{sec/0_abstract}  
\input{sec/introduction}

\input{sec/related_work}
\input{sec/materials_and_methods}

\input{sec/exp_setup}

\input{sec/results}
\input{sec/conclusions}

\clearpage  


%
%
\bibliographystyle{splncs04}
\bibliography{main}

\end{document}

%% file: sec/0_abstract.tex
\begin{abstract}
The rapid rise of generative models has yielded synthetic images of striking realism, blurring the line between real and fake content. 
As novel models proliferate, detectors must go beyond mere fake identification to robustly generalise across unseen generators and synthetic content.
We introduce FRIDA (Fake image Recognition and source Identification via Diffusion features Analysis), a lightweight, data-efficient framework that uses features from a pre-trained Stable Diffusion Model to detect and attribute AI-generated images.
Through an in-depth analysis of how data from different generators are encoded across diffusion U-Net layers, we propose a method that (i) detects synthetic images using a training-free $k$-Nearest Neighbour approach and (ii) performs source model attribution via a compact neural classifier.
On the GenImage benchmark, FRIDA achieves state-of-the-art cross-generator detection with limited data while maintaining robust source model attribution capabilities. 
These results establish diffusion features as a reliable framework for AI-generated image forensics. 
\end{abstract}

%% file: sec/introduction.tex
\section{Introduction}
The outstanding capabilities of recent generative models to synthesize media with exceptional fidelity and variety~\cite{midjourney,nanobanana,betker2023improving,Rombach_2022_CVPR} have accelerated the proliferation of AI‑generated content across domains, including medical imaging~\cite{KAZEROUNI2023102846,WANG2025102593,10937272}, education~\cite{xia2024scoping,wu2025systematic}, marketing~\cite{grewal2025generative,cillo2025generative}, and robotics~\cite{wolf2025diffusion}.
This rapid progress has been largely driven by diffusion-based generators---including Stable Diffusion Model (SDM)~\cite{Rombach_2022_CVPR}, DALL-E  ~\cite{ramesh2021zero}, and Imagen~\cite{saharia2022photorealistic}---which produce highly realistic and semantically coherent visuals from simple text prompts.
This increasing accessibility of advanced image generation tools through user-friendly interfaces poses considerable risks, including the creation of biased content and the violation of intellectual property rights.
To mitigate these risks, robust detection and attribution methods are required to reliably identify synthetic media and determine their source models~\cite{sha2023fake, luo2024lare, cioni2024clip}.
Early fake detection methods relied on supervised learning pipelines~\cite{wu2025explainable,zhu2023genimage,luo2024lare}, which depend on vast labelled datasets and intensive computation.
Approaches for source model attribution pushed these requirements even further, as they must learn generator-specific fingerprints~\cite{yu2019attributing,bui2022repmix,wang2020cnn}.
However, as generative models evolve rapidly, the continual need for data collection and retraining makes classifier-based detectors hard to scale, underscoring the need for lightweight, data-efficient methods that generalize across both known and emerging generators.

An opportunity to move beyond task-specific classifiers has emerged with the rise of large-scale foundation models based on Vision Transformers (ViTs)~\cite{dosovitskiy2021an}, e.g., CLIP~\cite{radford2021learning} and DINO~\cite{caron2021emerging}.
Pre-trained on massive datasets, these models learn general, transferable representations that have shown remarkable effectiveness in downstream tasks~\cite{kerssies2025your,wang2024enhancing}---including deepfake detection~\cite{nguyen2024exploring,smeu2025declip,wang2025fsfm}---when their extracted features are used for classification.

Similarly, diffusion-based generative models, though designed for image generation, have proven to be remarkably effective feature encoders.
Recent studies have shown that pre-trained SDMs provide rich semantic information about image content~\cite{BONECHI20245244,BONECHI2025128846}, enabling strong performance in applications such as semantic segmentation~\cite{baranchuk2021label,corradini2024freeseg}, and other vision-related problems---substantially reducing the computational cost of training large models from scratch. 
At the same time, diffusion models have been increasingly used for fake detection.
Some methods identify diffusion-generated images through reconstruction error~\cite{wang2023dire}, while others analyse latent trajectories via diffusion inversion to enhance generalisation across unseen generators~\cite{cazenavette2024fakeinversion,vasilcoiu2025latte}.
Based on the principle that images from different sources exhibit distinct distributions, Zhong et al.~\cite{zhong2025beyond} employ a SDM as a denoising tool to identify fake images by detecting the resulting artifacts in their reconstructed versions.
However, these approaches typically require repeated diffusion or inversion steps, making them computationally expensive and less practical for large-scale or real-time detection scenarios.

Inspired by these recent advances, we explore whether the internal features of a pre-trained diffusion-based generative model can effectively separate real from synthetic images drawn by various generators and, beyond detection, enable attribution to the source generator.
To this end, we propose FRIDA, a data-efficient framework for detecting fake images. 
The core of our method is to repurpose a pre-trained SDM (v1.5) at the last inference step as a feature extractor. 
Then, these features are employed by a simple $k$-Nearest Neighbours ($k$-NN) classifier, showing state-of-the-art results on the GenImage benchmark~\cite{zhu2023genimage}.
This strategy eliminates the need for costly inversion or fine-tuning and shows strong generalisation to unseen generators.
We further extend our method to source model attribution, showing that a lightweight Multi-Layer Perceptron (MLP) trained on the SDM latent features can accurately identify the generator responsible for a synthetic image.

The main contributions of our work can be summarized as follows:
\begin{enumerate}
    \item We show that latent features extracted from specific layers of a pre-trained SDM are highly discriminative for fake image detection. 
    Furthermore, we benchmark the efficacy of SDM as a feature extractor against prominent ViT-based backbones (CLIP and DINO), establishing that SDM-based representations achieve superior detection performance.
    \item We evaluate our fake image detection approach on data generated by eight different models, achieving state-of-the-art performance on GenImage dataset with a novel $k$-NN-based framework that is entirely training-free. 
    We further demonstrate the robustness of our method by testing on a challenging unseen subset comprising state-of-the-art generators, including Flux and SDv3.5.
    Our method requires only a small support set, generalises effectively to unseen generators, and adapts to new data without any fine-tuning.
    \item We evaluate the robustness of FRIDA against varying levels of image perturbations, including Gaussian noise and JPEG compression, to simulate real-world distortions.
    \item A lightweight neural classifier trained on the SDM latent representation proves to be highly effective in source model attribution, suggesting the presence of generator-specific characteristics in the features. 
    We use SHAP (SHapley Additive exPlanations)~\cite{lundberg2017unified} to further investigate the capability of the SDM to encode these model-specific signatures.
\end{enumerate}

\noindent The remainder of this paper is organized as follows: ~\Cref{sec:rel_work} reviews relevant literature on synthetic image detection and attribution. 
\Cref{sec:materials_and_methods} presents the benchmark dataset (GenImage) alongside an auxiliary dataset of recent generators, and outlines the methodology for extracting per-image latent representations using the SDM. 
~\Cref{sec:exp_setup} describes the experimental setup, while ~\Cref{sec:results} presents and discusses the obtained results. 
Finally, ~\Cref{sec:conclusions} concludes the paper with a summary of our findings and outlines future research directions.

%% file: sec/related_work.tex
\section{Related Work}
\label{sec:rel_work}

\paragraph{Frequency- and Texture-based Detectors.}
Early research on detecting GAN-generated content showed that synthetic images contain characteristic frequency patterns and texture artifacts.
Zhang et al.~\cite{zhang2019detecting} introduced AutoGAN, a frequency-domain simulator of upsampling artifacts, and trained a spectrum-based classifier to detect periodic Fourier patterns without access to the generator.
Similarly, Wang et al.~\cite{wang2020cnn} showed that CNN-generated images share common low-level artifacts across architectures and datasets.
Their ResNet-based detector, trained on images from ProGAN, generalized remarkably well to unseen generators such as StyleGAN and BigGAN, revealing universal CNN fingerprints that persist across models.
Gram-Net~\cite{liu2020global} models global texture correlations via multi-layer Gram matrices, improving robustness to compression and noise and enabling better cross-GAN generalization by leveraging long-range texture cues.
Qian et al.~\cite{qian2020thinking} advanced this line of work with F$^3$-Net, a two-branch frequency-aware CNN classifier combining frequency decomposition and local statistics via cross-attention, outperforming spatial-domain detectors on a deepfake faces dataset even under strong compression.

\paragraph{Diffusion-based Detection and Reconstruction Methods.}
Following the success of diffusion-based models, several works have focused on detecting images generated by such models.
DE-FAKE~\cite{defake2023} is a hybrid two-branch architecture that exploits image-text consistency between captions and visual content to detect synthetic images.
DIRE~\cite{wang2023dire} detects diffusion-generated images by inverting an input into the diffusion latent space and then reconstructing it through the full reverse denoising trajectory; the discrepancy between the original and reconstructed signals is used for detection.
Similarly, LaRE$^2$~\cite{luo2024lare} estimates a Latent Reconstruction Error using only a single-step reconstruction in latent space, using a module that aligns and refines features across spatial and channel dimensions.
A related line of work~\cite{diffimplicit2025} also leverages a pre-trained diffusion model, extracting multi-timestep responses under the hypothesis that synthetic images, which lie outside the natural image manifold, exhibit distinctive denoising behavior.
ESIDE~\cite{wu2025explainable} incorporates frequency perturbations into diffusion inversion, training an ensemble of CLIP-based classifiers on noised representations to improve robustness and interpretability. 
More recently, LATTE~\cite{vasilcoiu2025latte} introduces a Latent Trajectory Embedding framework that explicitly models the temporal evolution of latent representations across denoising steps.
By aggregating multi-timestep latent features through joint visual-latent refinement, LATTE captures dynamic generation cues beyond reconstruction error, achieving strong cross-generator and cross-dataset generalization.
An alternative approach~\cite{zhong2025beyond} treats detection as an anomaly detection task. It learns the low-level feature distribution of real images by training an extractor to spot pixel-level differences between original images and their denoised counterparts, effectively identifying generated content that falls outside this learned distribution.

\paragraph{Semantic and Open-Set Attribution Frameworks.}
To generalize across diverse generators, recent research explores semantic alignment, which disentangles generator-specific artifacts from image semantics, and open-set detection, which extends attribution to generators unseen at training time.
Zhu et al.~\cite{zhu2025maid} propose MAID, a framework-agnostic attribution method that extracts diffusion model activations by treating pre-trained diffusion models as denoising autoencoders. 
These activations encode model-specific patterns without requiring white-box access or prompts, supporting both detection and attribution. 
SemGIR~\cite{semgir2024} employs semantic-guided image regeneration: a candidate image is captioned, regenerated via text-to-image synthesis, and compared with its reconstruction using CLIP-based encoders. 
This forces the detector to focus on generator-specific artifacts rather than prompt semantics and achieves strong cross-generator generalization. 
Cioni et al.~\cite{clipfeat2024} move beyond frequency-based fingerprints by leveraging intermediate representations of large ViT-based models such as CLIP and DINO. 
Their open-set attribution framework combines linear probing and $k$-nearest neighbors with confidence-based or distance-based rejection to identify images from unseen generators, providing a unified and retraining-free solution across GAN and diffusion sources.

%% file: sec/materials_and_methods.tex
\section{Dataset Overview and Image Prototyping}
\label{sec:materials_and_methods}

\subsection{Datasets}
\label{sec:datasets}

\paragraph{GenImage Dataset.} \label{sec:genimage_dataset}
GenImage~\cite{zhu2023genimage} is a large-scale benchmark designed to detect generated imagery, composed of both synthetic and real images sourced from ImageNet~\cite{deng2009imagenet}. 
The dataset contains roughly 1.35 million synthetic images produced by eight distinct generative models: BigGAN~\cite{brock2018large}, GLIDE~\cite{nichol2022glide}, VQDM~\cite{gu2022vector}, SDM (v1.4 and v1.5)~\cite{Rombach_2022_CVPR}, ADM~\cite{dhariwal2021diffusion}, Midjourney~\cite{midjourney}, and Wukong~\cite{wukong}. Both the real and synthetic images cover the 1,000 ImageNet classes, with the synthetic portion providing 1,350 images per class (1,300 for training and 50 for testing). 
The generated images were created using simple text prompts following the template ``photo of [class]'' and have resolutions ranging from $128\times128$ to $1024\times1024$, depending on the source model.
In this study, we utilize a subset of the training images. 
For each generator, we randomly sampled 10,000 real and 10,000 synthetic images (10 images per ImageNet class). 
These subsets are then partitioned into training ($80\%$) and validation ($20\%$) sets. 
In the rest of the paper, we will refer to these two sets as ``training subset'' and ``validation set''.
Finally, the original GenImage test set was used as a held-out evaluation set.

\paragraph{Other Datasets.}
\label{sec:other_datasets}
To evaluate generalization beyond GenImage, we assembled an alternative, out-of-distribution (OOD) test set by extracting 500 synthetic images per generator and 500 real images per dataset from different sources.
In particular, fake images are sampled from the Qwen-Image-Self-Generated-Dataset~\cite{wu2025qwenimagetechnicalreport} and from five generators of the OpenFake dataset~\cite{livernoche2025openfake} (i.e., Flux 1.1 Pro, Flux 1.0 Dev, GPT Image 1, Ideogram 3.0, and SDv3.5). 
The real images are sampled from LAION 400M~\cite{schuhmann2021laion} and Pascal VOC~\cite{Everingham15}.

\subsection{The GenImage Evaluation Protocol}
\label{sec:genimage_protocol}
The cross-generator evaluation protocol used in this work follows the one in ~\cite{zhu2023genimage}. 
To assess generalization, a binary real-vs-fake classifier is trained on one generator-specific subset and then tested---without any adaptation---on evaluation data from both the same generator (seen data distribution) and from the other seven generators (unseen data distributions). 

\subsection{Image Prototype Extraction}
\label{sec:prototype_extraction}
\begin{figure}[ht!]
\vspace{-0.4cm}
  \centering
  \includegraphics[width=0.6\columnwidth]{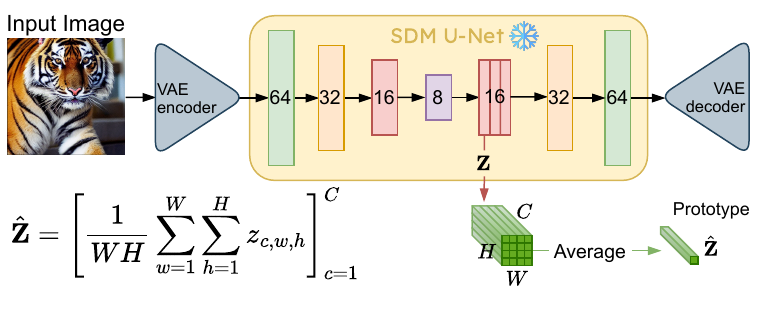}
  \caption{\textbf{Prototype extraction from Stable Diffusion U-Net.} In this example, we extract and average the features from the first decoder layer at $16\times16$ resolution.}
  \label{fig:prototype_extraction}
\vspace{-0.4cm}
\end{figure}

We adopt a pre-trained SDMv1.5\footnote{\url{https://github.com/hkproj/pytorch-stable-diffusion}} and we use the following procedure for extracting internal features prototype from a specific layer (see~\Cref{fig:prototype_extraction}).

Each input image is resized to $512 \times 512$, and we run a forward pass at $t=0$, which corresponds to the final denoising step of the diffusion U-Net.
After encoding the image into a latent representation via the VAE, this latent is passed to the U-Net. 
Compact prototypes are then obtained by spatially averaging the feature maps extracted from a chosen U-Net layer.


%% file: sec/exp_setup.tex
\section{Experimental Setup}
\label{sec:exp_setup}
This section details the experimental setup used in our study.
First, we compare features from different layers of the diffusion U-Net to identify which layer is most effective at distinguishing real from synthetic content (\Cref{sec:layer_sel}).
We then describe the experimental setups for our two primary tasks: fake image detection (\Cref{sec:fake_detection}) and source model attribution (\Cref{sec:attribution}).
Finally, in~\Cref{subsec:sdm_feature_ablations}, we present additional studies using different backbones, out-of-distribution datasets and various image perturbations.


\subsection{U-Net Layer Selection by Linear Probing}
\label{sec:layer_sel}
To identify which layers provide the most effective features for fake image detection, we extract latent representations from the encoder, bottleneck, and decoder layers of the U-Net at multiple spatial resolutions ($64\times64$, $32\times32$, $16\times16$, and $8\times8$).
We adopt a linear probing setup---commonly used to evaluate the quality and separability of learned representations in self-supervised and foundation models ~\cite{chen2020simple,he2020momentum,radford2021learning}---by training a linear classifier with sigmoid activation on the latent prototypes from each layer.
Each classifier is trained on the training subset (see~\Cref{sec:genimage_dataset}) with real and synthetic images from a given generator. 
All models are trained with the AdamW optimizer~\cite{loshchilov2018decoupled} using a learning rate of $3\times10^{-4}$. Training is early stopped based on validation accuracy, with a patience of 15 epochs. 
To assess generalization, we perform the cross-generator evaluation described in \Cref{sec:genimage_protocol} using the eight validation sets.
The U-Net layer whose features yield the highest average cross-generator validation accuracy is then selected and used for all subsequent experiments.

\subsection{Fake Image Detection}
\label{sec:fake_detection}
Our first goal is to design a fake image detection pipeline capable of generalizing to images from unseen generators.
To this aim, we evaluate two approaches: a simple neural network (MLP) and a distance-based classifier ($k$-NN). In both cases, we extract the features from the selected U-Net layer and  follow the cross-generator evaluation protocol described in~\Cref{sec:genimage_protocol} to select the best model configurations based on the validation accuracy.
The best MLP and $k$-NN models are then tested on the GenImage test set.

\paragraph{MLP for Fake Image Detection.}
We evaluate three MLPs with different numbers of hidden units and hidden layers. 
In particular, we consider the following configurations:
\begin{itemize}
    \item \texttt{\textbf{MLP-640}}: A single hidden layer with 640 units.
    \item \texttt{\textbf{MLP-320}}: A single hidden layer with 320 units.
    \item \texttt{\textbf{MLP-640-320}}: Two hidden layers with 640 and 320 units, respectively.
\end{itemize}
All the models have two output neurons with a softmax activation function and are trained using binary cross-entropy loss with AdamW optimizer and a learning rate of $3\times10^{-4}$. 
The validation accuracy is used to early stop the training process with a patience of 15 epochs.

\paragraph{$k$-NN for Fake Image Detection.}
For the $k$-NN approach, we extract a balanced support set $S_G$ ($50\%$ real, $50\%$ fake) for each generator $G$ from the training subset. 
Hyperparameters are tuned through an extensive search over different distance metrics (Euclidean, Correlation~\cite{szekely2007distance}, Manhattan~\cite{krause1973taxicab}, Cosine~\cite{manning2008introduction}),  $S_G$ sizes (ranging from 4 to 1,000)\footnote{$Sizes(S_{G})=$\{4, 10, 20, 40, 80, 100, 200, 300, 400, 600, 800, 1000\}}, and number of neighbours $k$ (from 1 to 45)\footnote{$k=$\{1, 3, 5, 7, 9, 11, 13, 15, 17, 19, 21, 23, 25, 27, 29, 31, 33, 35, 37, 39, 41, 43, 45\}}. 


\subsection{Source Model Attribution}
\label{sec:attribution}
Our second objective is image attribution, a multi-class classification task to identify the source model of an image.
As for fake image detection, we compare the MLP and $k$-NN approaches, adapting the models and the evaluation protocol for multi-class classification.
More specifically, we employ a single classifier with nine output classes (eight generators + real images).

\paragraph{MLP for Source Model Attribution.}
We evaluate the same MLP configurations used for fake-image detection (see~\Cref{sec:fake_detection}), but we replace the output layer with a 9-way softmax classifier and train with cross-entropy loss. 
All MLPs are trained on synthetic images from the training subsets of all generators, while the real images are sampled from the training subset to maintain the class balance. 
All models are trained using the AdamW optimizer with a learning rate of $3\times10^{-4}$, and the validation accuracy is used to early stop the training process using a patience of 15 epochs.

\paragraph{$k$-NN for Source Model Attribution.}
We perform the hyperparameter selection for $k$-NN using the same grid employed in~\Cref{sec:fake_detection}, changing the sizes of the support set $S$ (ranging from 18 to 9,000)\footnote{$Sizes(S)=$\{18, 45, 90, 180, 360, 450, 900, 1350, 1800, 2700, 3600, 4500, 9000\}}
to comprise an equal number of images for all nine classes.

\subsection{Extended Experimental Analysis}
\label{subsec:sdm_feature_ablations}

\paragraph{Benchmarking Against ViT Backbones.}
\label{subsubsec:abl_clip_dino}
We assess the advantage of using SDM features in our method by conducting a comparative analysis against established ViT-based feature extractors, specifically CLIP and DINO. 
For both, we extract the features from their last layer (widely used for their robust performance across various downstream tasks) and we evaluate the cross-generator performance under the same setup used for the SDM.


\paragraph{Robustness to Image Corruptions.}
\label{subsubsec:abl_perturbations}
We evaluate the robustness of FRIDA by introducing various synthetic distortions to the test set. 
Specifically, we assess the cross-generator performance across multiple degradation levels, including JPEG compression (10, 30, 50, 70, 90), gaussian noise ($\sigma$=0.05, 0.1, 0.15, 0.2, 0.25), and gaussian blur ($\sigma$=1, 2, 3, 4, 5).

\paragraph{Out-of-Distribution Robustness.}
\label{subsubsec:abl_other_generators}
We further evaluate the generalization capabilities of our approach on OOD datasets. 
The support set for $k$-NN is a random subset of the GenImage dataset encompassing all available generators and their corresponding real images. 
During inference, we evaluate the model against a series of unseen OOD generators. 
To isolate the domain shift introduced by the synthetic data, the real images in the test set remain constant across all evaluations, comprising 500 samples balanced equally between Pascal VOC ($N=250$) and LAION-400M ($N=250$). 
For each test iteration, these are paired with 500 synthetic images generated by a single OOD model, thereby maintaining a strictly balanced binary classification setup.

%% file: sec/results.tex
\section{Results}
\label{sec:results}
This section presents our experimental results, starting with the U-Net layer selection for feature extraction (\Cref{sec:res_layer_sel}), and then covering fake image detection (\Cref{sec:res_fake_detection}) and source model attribution (\Cref{sec:res_image_attribution}).
Finally, in~\Cref{subsec:extended_results}, we present the results of the extended analysis using different backbones, OOD datasets and various image perturbations.

\subsection{U-Net Layer Selection by Linear Probing}
\label{sec:res_layer_sel}
\begin{figure}[ht!]
\vspace{-0.4cm}
  \centering
  \includegraphics[width=0.48\columnwidth]{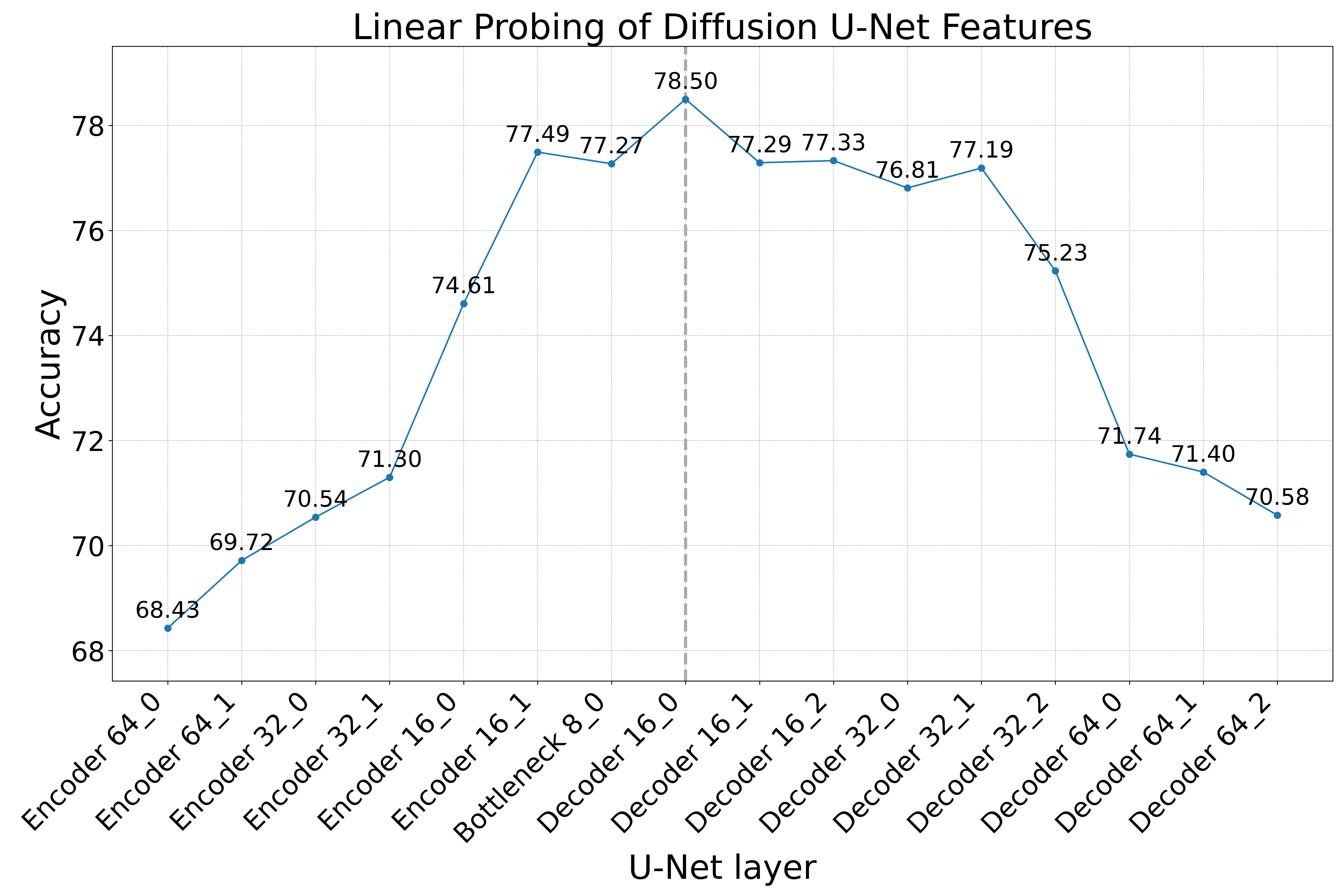}
  \caption{\textbf{U-Net Layer Selection by Linear Probing.} 
  Average cross-generator validation accuracy obtained using the prototypes extracted from different U-Net layers (labelled as Encoder, Bottleneck or Decoder, followed by the spatial resolution and the intra-stage index). 
  The best accuracy is achieved by features from the first layer of the decoder at $16\times16$ resolution.}
  \label{fig:res_layer_sel}
  \vspace{-0.4cm}
\end{figure}

Following the experimental setup in~\Cref{sec:layer_sel}, we test image prototypes from all the layers of the diffusion U-Net using a linear probing approach to determine (i) if the latent space representation can be used to distinguish between real and synthetic data, and (ii) which layer produces the most informative representation for this task.
\Cref{fig:res_layer_sel} plots the average cross-generator validation accuracy for features extracted from different levels of the diffusion U-Net.

The results of our feature probing analysis show that the latent features extracted from the diffusion U-Net can be highly discriminative for distinguishing real from synthetic images.
In particular, we identify the first layer of the decoder at $16\times16$ resolution (\texttt{\textbf{Decoder 16\_0}}) as the source of the most informative features, yielding a peak average cross-generator validation accuracy of $78.50\%$. 
Given this result, we adopt these features in the remainder of our study.

\subsection{Fake Image Detection}
\label{sec:res_fake_detection}
Following the experimental setup described in~\Cref{sec:fake_detection}, we evaluate two methods: one based on neural networks (i.e., MLP) and the other on a $k$-NN approach.

\paragraph{Neural Network Classifier.}
We conduct preliminary experiments using the three MLPs described in~\Cref{sec:fake_detection}.
MLP-320, which achieves an average cross-generator accuracy of $79.1\%$, outperforms MLP-640 ($78.2\%$) and MLP-640-320 ($78.9\%$). 
A detailed breakdown of the results for MLP-320 on cross-generator fake image detection is presented in~\Cref{tab:mlp-360_res_val}.

\begin{table}[ht!]

\centering
\scriptsize
\setlength{\tabcolsep}{1.9pt} 
\resizebox{0.7\textwidth}{!}{
\begin{tabular}{@{}l|cccccccc|c@{}}
\hline
\multirow{2}{*}{\shortstack{\textbf{Trained}\\\textbf{on}}} & \multicolumn{9}{c@{}}{\textbf{Accuracy (\%) when tested on}} \\
 & Midj. & SDv1.4 & SDv1.5 & ADM & Glide & Wukong & VQDM & BigGAN & \textbf{Avg.} \\
\hline
Midj.    & 98.0 & 85.6 & 84.6 & 68.0 & 89.9 & 79.7 & 52.1 & 49.8 & 75.9 \\
SDv1.4   & 89.2 & 98.8 & 98.5 & 61.7 & 76.3 & 94.5 & 61.2 & 49.1 & 78.6 \\
SDv1.5   & 87.5 & 98.8 & 98.8 & 59.1 & 70.9 & 94.8 & 60.5 & 49.3 & 77.5 \\
ADM      & 67.6 & 74.2 & 75.5 & 99.3 & 98.8 & 68.1 & 88.3 & 93.9 & 83.2 \\
Glide    & 63.8 & 64.7 & 64.9 & 94.4 & 99.7 & 59.1 & 85.3 & 87.7 & 77.5 \\
Wukong   & 88.3 & 98.0 & 98.2 & 86.6 & 93.6 & 98.1 & 86.6 & 59.2 & 88.6 \\
VQDM     & 57.6 & 64.4 & 64.7 & 94.9 & 99.5 & 61.4 & 99.7 & 99.5 & 80.2 \\
BigGAN   & 54.0 & 53.0 & 53.0 & 79.8 & 99.4 & 53.1 & 80.2 & 99.8 & 71.5 \\
\hline
\multicolumn{1}{@{}l|}{\textbf{Avg.}} & 75.7 & 79.7 & 79.8 & 80.4 & 91.0 & 76.1 & 76.7 & 73.5 & 79.1 \\
\hline
\end{tabular}
}
\caption{\textbf{Cross-generator evaluation of the MLP-320 for fake image detection.} We train the model on each generator and test it on the validation set of all the generators.}
\label{tab:mlp-360_res_val}
\vspace{-0.8cm}
\end{table}

The results reveal a clear pattern: while the MLP effectively learns generator-specific artifacts (as evidenced by the high diagonal accuracy), it struggles to generalize to unseen generators.
We can also observe the presence of distinct generator families that share underlying characteristics within the latent space of the diffusion U-Net.
For example, an MLP trained on SDM latents of Midjourney images transfers well to SDv1.4, SDv1.5, and Glide, suggesting a strong similarity in their latent feature representations.

\paragraph{$k$-NN Classifier.} 
Although U-Net features effectively distinguish real from synthetic data, the MLP classifier tends to learn generator-specific artifacts, hindering generalization.
To address this limitation, we propose a training-free $k$-NN approach that operates on distances between feature prototypes. 
By relying on the intrinsic geometry of the feature space rather than learned parameters, $k$-NN leverages the structure already encoded in the representations, leading to more effective 
cross-generator generalization.
To this aim, guided by the hyperparameter search described in~\Cref{sec:fake_detection}, we select the $k$-NN model using correlation distance with $k=45$ and a size of $S_G=1000$. 
This configuration achieves the best validation accuracy, $85.3\%$ across eight generators as shown in~\Cref{tab:best_knn_res_val}, and surpasses the best MLP ($79.1\%$). 
The same $k$-NN configuration is then evaluated on the GenImage test set (\Cref{tab:best_knn_res_test}).

\begin{table}[ht!]
\vspace{-0.2cm}
\centering
\scriptsize
\setlength{\tabcolsep}{2.2pt}

\begin{subtable}[t]{0.49\columnwidth}
\centering
\caption{Results on GenImage Validation set}
\begin{adjustbox}{width=\linewidth}
\begin{tabular}{@{}l|cccccccc|c@{}}
\hline
\multirow{2}{*}{\textbf{$S_G$}} & \multicolumn{9}{c@{}}{\textbf{Accuracy (\%) when tested on}} \\
 & Midj. & SDv1.4 & SDv1.5 & ADM & Glide & Wukong & VQDM & BigGAN & \textbf{Avg.} \\
\hline
Midj.  & 88.0 & 85.8 & 86.3 & 83.8 & 87.9 & 87.6 & 86.1 & 85.1 & 86.3 \\
SDv1.4 & 88.1 & 90.0 & 90.7 & 83.5 & 87.7 & 65.9 & 84.7 & 88.1 & 84.8 \\
SDv1.5 & 89.0 & 89.2 & 89.7 & 83.0 & 88.6 & 85.5 & 86.6 & 88.2 & 87.5 \\
ADM    & 83.5 & 82.4 & 83.6 & 82.6 & 84.0 & 84.2 & 84.5 & 82.6 & 83.4 \\
Glide  & 85.9 & 84.4 & 84.5 & 85.7 & 87.8 & 87.8 & 88.5 & 84.3 & 86.1 \\
Wukong & 82.1 & 74.8 & 75.7 & 84.5 & 89.7 & 89.0 & 88.5 & 76.8 & 82.6 \\
VQDM   & 82.7 & 83.2 & 83.8 & 83.9 & 85.1 & 85.2 & 85.9 & 83.6 & 84.2 \\
BigGAN & 86.9 & 87.3 & 88.3 & 83.4 & 88.3 & 88.6 & 88.4 & 87.3 & 87.3 \\
\hline
\textbf{Avg.} & 85.8 & 84.6 & 85.3 & 83.8 & 87.4 & 84.2 & 86.6 & 84.5 & 85.3 \\
\hline
\end{tabular}
\end{adjustbox}
\label{tab:best_knn_res_val}
\end{subtable}
\hfill
\begin{subtable}[t]{0.49\columnwidth}
\centering
\caption{Results on GenImage Test set}
\begin{adjustbox}{width=\linewidth}
\begin{tabular}{@{}l|cccccccc|c@{}}
\hline
\multirow{2}{*}{\textbf{$S_G$}} & \multicolumn{9}{c@{}}{\textbf{Accuracy (\%) when tested on}} \\
 & Midj. & SDv1.4 & SDv1.5 & ADM & Glide & Wukong & VQDM & BigGAN & \textbf{Avg.} \\
\hline
Midj.  & 91.1 & 88.4 & 88.7 & 87.3 & 91.0 & 90.8 & 88.8 & 88.6 & 89.3 \\
SDv1.4 & 89.9 & 91.5 & 91.1 & 85.3 & 88.4 & 67.7 & 85.2 & 89.8 & 86.1 \\
SDv1.5 & 91.6 & 91.2 & 91.1 & 85.8 & 91.0 & 87.9 & 88.8 & 90.3 & 89.7 \\
ADM    & 86.4 & 85.7 & 85.9 & 87.0 & 87.1 & 86.8 & 86.9 & 85.2 & 86.4 \\
Glide  & 89.3 & 87.0 & 86.9 & 89.8 & 91.4 & 90.8 & 90.4 & 87.5 & 89.1 \\
Wukong & 85.4 & 78.1 & 78.3 & 88.0 & 93.4 & 93.1 & 91.3 & 80.8 & 86.1 \\
VQDM   & 86.9 & 86.7 & 86.7 & 88.3 & 89.0 & 88.6 & 88.5 & 86.1 & 87.6 \\
BigGAN & 89.8 & 89.6 & 89.5 & 86.3 & 90.5 & 91.5 & 90.3 & 89.5 & 89.6 \\
\hline
\textbf{Avg.} & 88.8 & 87.3 & 87.3 & 87.2 & 90.2 & 87.2 & 88.8 & 87.2 & 88.0 \\
\hline
\end{tabular}
\end{adjustbox}
\label{tab:best_knn_res_test}
\end{subtable}

\caption{\textbf{Cross-generator evaluation of $k$-NN on the GenImage dataset for fake image detection.} 
Accuracy of $k$-NN ($k=45$) classifier with support set size of 1,000 using distance correlation.}
\label{tab:knn_combined}
\end{table}

\begin{table}[ht!]
\centering

\normalsize
\resizebox{0.7\textwidth}{!}{
\begin{tabular}{@{}lccccccccc@{}}
\toprule
\textbf{Methods} & \multicolumn{8}{c}{\textbf{Tested on}} & \textbf{Avg. } \\
\cmidrule(lr){2-9}
& Midjourney & SDv1.4 & SDv1.5 & ADM & GLIDE & Wukong & VQDM & BigGAN & \\
\midrule
CNNSpot \cite{wang2020cnn} & 58.2 & 70.3 & 70.2 & 57.0 & 57.1 & 67.7 & 56.7 & 56.6 & 61.7 \\
Spec \cite{zhang2019detecting}    & 56.7 & 72.4 & 72.3 & 57.9 & 65.4 & 70.3 & 61.7 & 64.3 & 65.1 \\
F$^3$-Net \cite{qian2020thinking}   & 55.1 & 73.1 & 73.1 & 66.5 & 57.8 & 72.3 & 62.1 & 56.5 & 64.6 \\
GramNet \cite{liu2020global} & 58.1 & 72.8 & 72.7 & 58.7 & 65.3 & 71.3 & 57.8 & 61.2 & 64.7 \\
DIRE \cite{wang2023dire}\space\dag & 65.0 & 73.7 & 73.7 & 61.9 & 69.1 & 74.3 & 63.4 & 56.7 & 67.2 \\
LaRE$^2$ \cite{luo2024lare}        & 66.4 & \underline{\textbf{87.3}} & \underline{87.1} & 66.7 & 81.3 & 85.5 & 84.4 & 74.0 & 79.1 \\
LATTE \cite{vasilcoiu2025latte} & 71.3 & 79.3 & 81.8 & 82.2 & \textbf{92.8} & 82.0 & 82.9 & \textbf{87.8} & {82.5} \\
\hline
\rowcolor{yellow!15} \multicolumn{1}{@{}l}{\textbf{FRIDA-200 (Ours)}} & \underline{86.9} & {86.2} & {86.1} & \underline{85.0} & {87.9} & \textbf{87.5} & \underline{86.9} & {85.9} & \underline{86.5} \\
\rowcolor{yellow!15} \multicolumn{1}{@{}l}{\textbf{FRIDA (Ours)}} & \textbf{88.8} & \underline{\textbf{87.3}} & \textbf{87.3} & \textbf{87.2} & \underline{90.2} & \underline{87.2} & \textbf{88.8} & \underline{87.2} & \textbf{88.0} \\
\bottomrule
\end{tabular}
}
\caption{\textbf{Comparison with state-of-the-art methods for fake image detection on the GenImage test set.} 
Given a specified test set, we evaluate eight $k$-NN classifiers---each based on the support set from one generator---and compute their average accuracy.
FRIDA-200 exploits a support set of 200 samples, reduced by $80\%$ w.r.t. FRIDA.
Best in \textbf{bold}, second best \underline{underlined}.
\dag\space results taken by \cite{luo2024lare}.
}
\label{tab:sota_comparison_detection}
\end{table}
As detailed in~\Cref{tab:sota_comparison_detection}, our approach demonstrates a significant leap in performance over current state-of-the-art methods on the GenImage test set. The experimental setup is designed to rigorously test generalization: eight models are trained on limited data (1,000 samples) from a single generator each, but evaluated on images from all generators. In this challenging scenario, our method is the one that better generalizes across these unseen data distributions, establishing a new state-of-the-art performance by a margin of nearly six percentage points.

Beyond its accuracy, the primary advantage of FRIDA in fake image detection lies in its remarkable data and training efficiency. 
The strong results are achieved without any conventional training process: we use a pre-trained SDM, and the $k$-NN framework merely requires storing the support set. 
The method also maintains high performance even with a drastically smaller support size. 
When the support set is reduced by $80\%$ (200 samples, referred to as FRIDA-200 in~\Cref{tab:sota_comparison_detection}), the average cross-generator accuracy on the test set drops by only $1.5\%$ (from $88.0\%$ to $86.5\%$), still outperforming the previous state-of-the-art (\Cref{tab:sota_comparison_detection}).
In a landscape where new generative models continually emerge, this property is crucial for practical deployment, as it avoids data-intensive retraining cycles.

\subsection{Source Model Attribution}
\label{sec:res_image_attribution}
We compare MLP and $k$-NN classifiers on their ability to identify an image's source generator, following the experimental procedure described in~\Cref{sec:attribution}.

\paragraph{$k$-NN Classifier.}
We adapt the distance-based approach used for fake image detection for the task of source model attribution. 
A portion of the results from the hyperparameters selection on the validation set obtained using the correlation distance is presented in~\Cref{tab:best_knn_val_attribution}.
These results prove that the $k$-NN classifier is inadequate for effective source model attribution. 
Despite optimization (using a support set of 9,000 and $k=9$), its peak accuracy reached only $57.7\%$, a performance level far too low for practical application.




\paragraph{Neural Network Classifier.}
We train the three MLP classifiers as described in~\Cref{sec:attribution}; in~\Cref{tab:res_mlp_val_attribution} we report the average results of the MLPs on the validation set across 10 training runs. 

\begin{figure}[ht!]
\centering
\begin{minipage}{0.43\columnwidth}
  \centering
  \includegraphics[width=\linewidth]{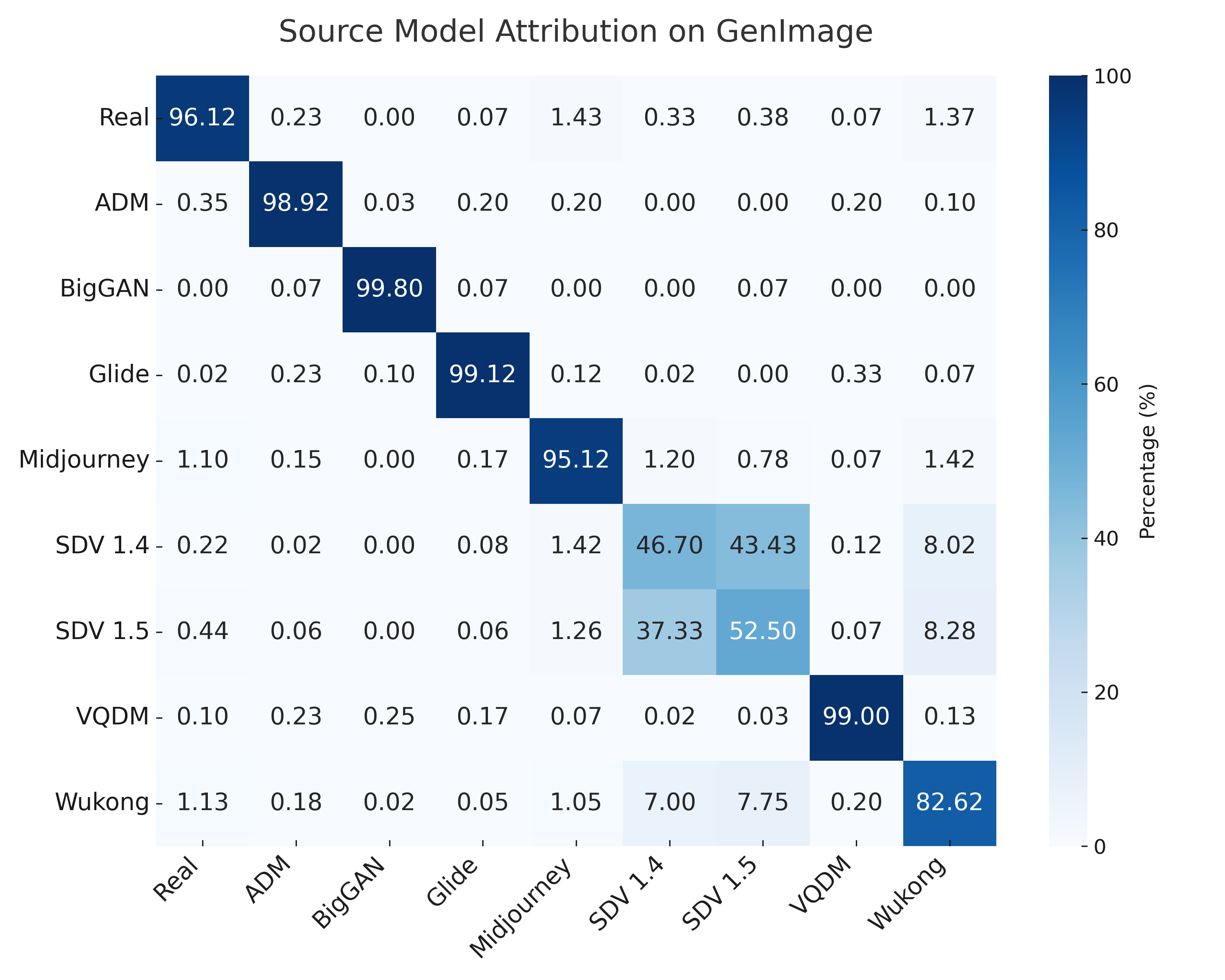}
  \captionof{figure}{\textbf{Confusion Matrix for source model attribution.} 
  The MLP-640 is evaluated on the GenImage test set.}
  \label{fig:conf_matrix_attribution}
\end{minipage}
\hfill
\begin{minipage}{0.43\columnwidth}
  \centering
  \includegraphics[width=\linewidth]{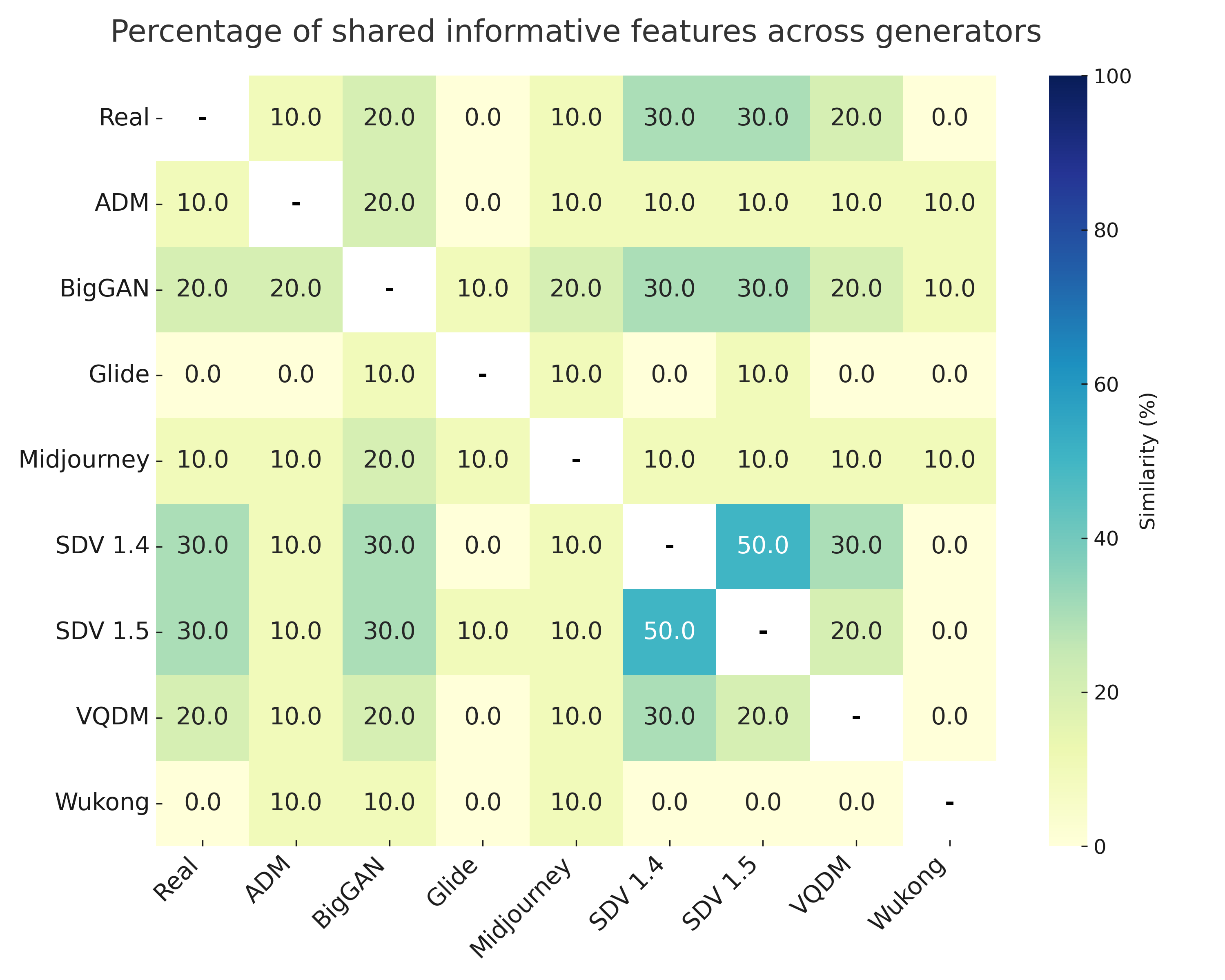}
  \captionof{figure}{\textbf{Shared informative features across generators.} 
  Percentage of overlap among the top 10 SHAP features.}
  \label{fig:perc_shared_features}
\end{minipage}
\vspace{-0.5cm}
\end{figure}

\begin{table}[h!]
\centering
\begin{minipage}{0.48\columnwidth}
\centering
\fontsize{7}{8}\selectfont
\begin{adjustbox}{width=\linewidth, center}
\begin{tabular}{@{}l|ccccccc}
\hline
\multirow{2}{*}{\textbf{$k$}} & \multicolumn{7}{c@{}}{\textbf{Support sizes}} \\
 & 18 & 90 & 180 & 900 & 1800 & 3600 & 9000 \\
\hline
1  & 25.2\% & 33.5\% & 36.4\% & 44.6\% & 48.1\% & 51.6\% & 55.8\% \\
9  & -- & 28.7\% & 31.3\% & 44.09\% & 48.7\% & 52.7\% & \textbf{57.7\%} \\
19 & -- & -- & 27.2\% & 42.5\% & 48.3\% & 52.1\% & 57.1\% \\
35 & -- & -- & -- & 39.5\% & 46.5\% & 50.9\% & 56.1\% \\
45 & -- & -- & -- & 37.8\% & 45.1\% & 50.2\% & 55.4\% \\
\hline
\end{tabular}
\end{adjustbox}
\captionof{table}{\textbf{Effect of $k$ and support set size on source model attribution accuracy.}
Validation accuracies for the nine-class attribution task.}
\label{tab:best_knn_val_attribution}
\end{minipage}
\hfill
\begin{minipage}{0.48\columnwidth}
\centering
\vspace{1.5em}
\fontsize{6}{7.2}\selectfont
\begin{tabular}{@{}l|cc}
\hline
\textbf{Model} & \textbf{Avg. Accuracy} & \textbf{Std.} \\
\hline
MLP-640     & \textbf{84.87\%} & 0.143 \\
MLP-320     & 84.71\% & 0.205 \\
MLP-640-320 & 84.78\% & 0.127 \\
\hline
\end{tabular}
\vspace{1.2em}
\captionof{table}{\textbf{Source model attribution performance of MLPs.}
Average validation accuracy and standard deviation over 10 runs.}
\label{tab:res_mlp_val_attribution}
\end{minipage}
\end{table}
While the performance of the three models on the source model attribution task are comparable, the MLP-640, with $84.87\%$ of average accuracy, slightly outperforms the others and allows for an increase in performance of about 27 percentage points over the $k$-NN classifier.
The MLP-640, when tested on the GenImage test set, achieves an accuracy of $84.36\%$. 
This result is consistent with the validation performance, indicating a high degree of generalization. 
As shown in~\Cref{fig:conf_matrix_attribution}, the classifier generally recognizes source generators with high accuracy; however, it struggles to distinguish models that share the same architecture. 
Specifically, the model tends to confuse SDv1.4 with SDv1.5 and vice versa, while occasionally misattributing the SDM-based Wukong model to the same group. 
We attribute this phenomenon to their common architectural family. 
To validate this hypothesis, we conducted a family-based analysis by grouping data from these related sources, achieving a source model attribution accuracy of 96.67\%.

Notably, the huge performance increase of the MLP over the $k$-NN classifier on the validation set confirms that the features extracted by the diffusion U-Net contain generator-specific patterns. 
The MLP is capable of learning these distinctive characteristics, whereas the $k$-NN approach, which relies solely on feature distances, cannot recognize these subtle patterns.
Although $k$-NN proves effective for the broader task of distinguishing real from synthetic content, it is insufficient for the more complex task of source model attribution. 
These results suggest that, while real and synthetic images can be detected through distances in the feature space, the discriminative power in source model attribution stems from more complex feature patterns that need an MLP to be captured.
To investigate this hypothesis, we employ the SHAP algorithm to interpret the decisions of the MLP-640 classifier. 
We utilize the Gradient Explainer with a background dataset of 200 samples and 1,000 test samples. 
For each class, we identify the top 10 most informative features. In~\Cref{fig:perc_shared_features} we report the percentage of the top 10 features shared between each pair of generators.

The two SDMs exhibit a $50\%$ overlap in their top 10 most important features. 
Furthermore, these shared features influence the model's output identically (see~\Cref{fig:shap_sdv14} and~\Cref{fig:shap_sdv15} in the Supplementary Material), indicating that the images generated by these two models lead to a latent representation with common characteristics.
As a result, the images generated by the two models are nearly indistinguishable to MLP. 
In contrast, the classifier successfully discriminates real and BigGAN images from the SDMs by exploiting divergent feature behavior. 
Although these models share $30\%$ of their top 10 features with the SDMs, the impact of these common features is markedly different (\Cref{fig:shap_real} and~\Cref{fig:shap_BigGAN} in the Supplementary Material). 
Finally, the Wukong model presents an interesting paradox. 
Despite occasional misclassifications with the SDM-generated images, the SHAP analysis reveals zero feature overlap within the top 10 most influential features (\Cref{fig:shap_wukong} in the Supplementary Material). 
This indicates that the classifier's errors are not driven by the same high-impact features that define the SDM class. 
Instead, the confusion likely arises because a different set of features in Wukong images produces a combined effect that coincidentally mimics the characteristics of an SDM, leading the classifier to an incorrect conclusion.


\subsection{Extended Experimental Analysis}
\label{subsec:extended_results}

\paragraph{Backbone Comparison.}
In~\Cref{tab:backbone_comparison} we report the results of the comparison among SDM, CLIP, and DINO as feature extractors following the setup described in~\Cref{subsubsec:abl_clip_dino}.
Features from SDM consistently outperform features from the ViT encoders, validating the assumption that diffusion-based features are inherently more effective at isolating the discriminative cues necessary for distinguishing real from generated imagery.

\begin{figure}[h]
\vspace{-0.5cm}
    \centering
    \begin{subfigure}[b]{0.27\linewidth}
        \centering
        \includegraphics[width=\linewidth]{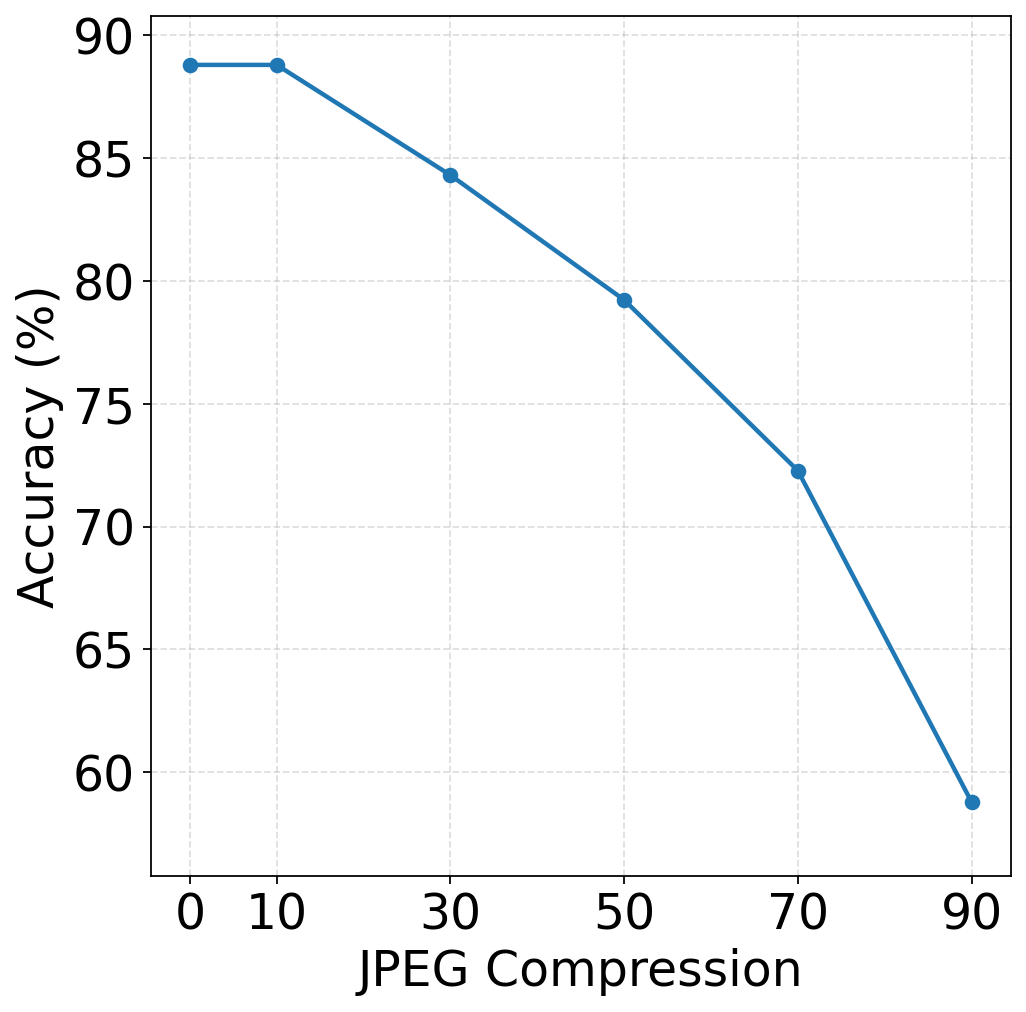}
        \caption{}
        \label{fig:jpeg_compression}
    \end{subfigure}
    \hfill
    \begin{subfigure}[b]{0.27\linewidth}
        \centering
        \includegraphics[width=\linewidth]{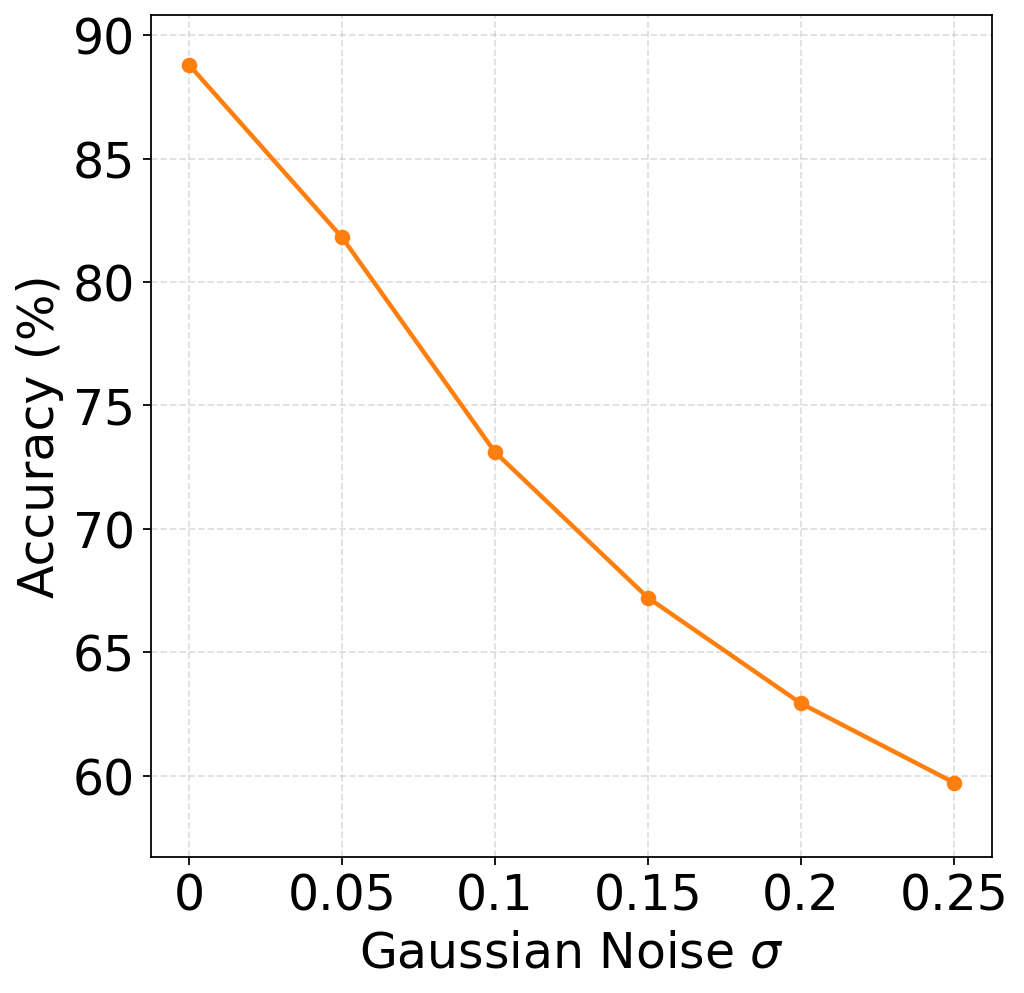}
        \caption{}
        \label{fig:noise}
    \end{subfigure}
    \hfill
    \begin{subfigure}[b]{0.27\linewidth}
        \centering
        \includegraphics[width=\linewidth]{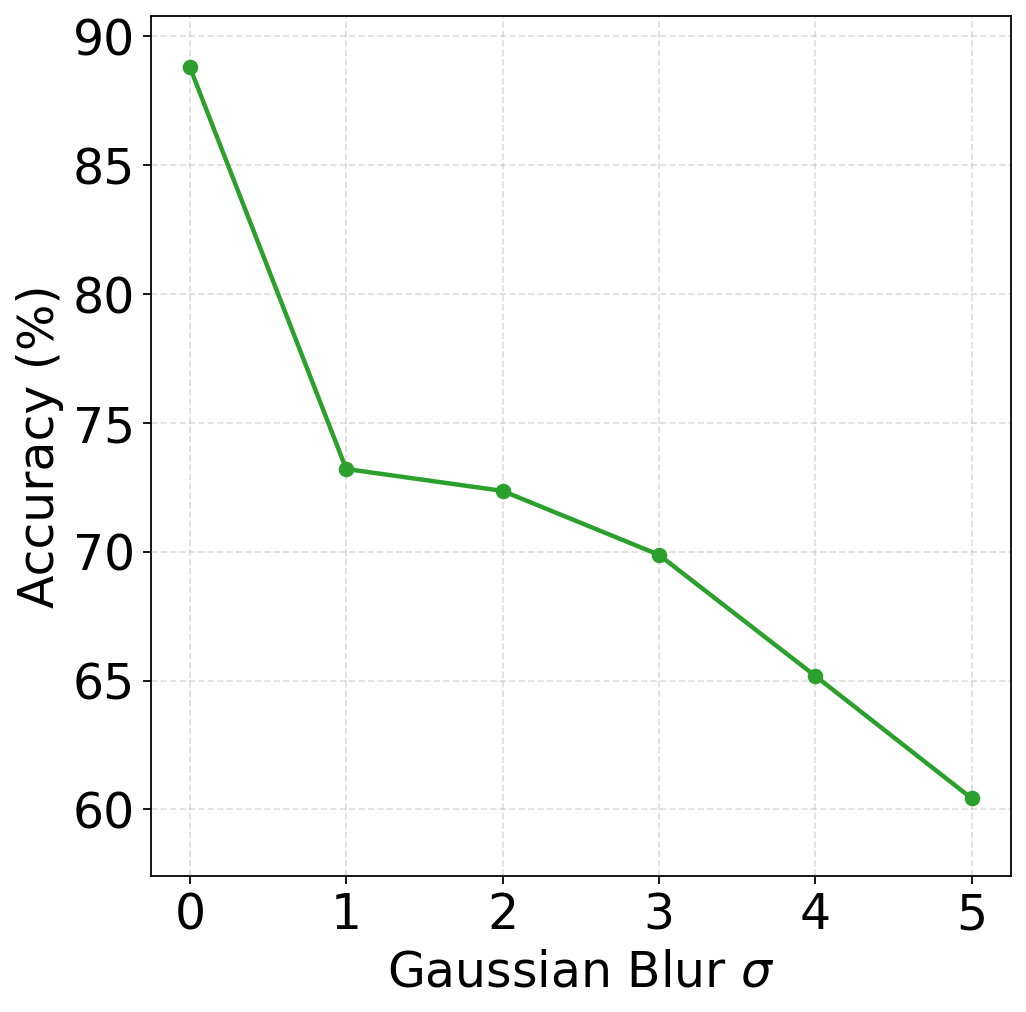}
        \caption{}
        \label{fig:blur}
    \end{subfigure}
    
    \caption{{\textbf{Robustness analysis against image distortions.} We report the detection accuracy for different image perturbations: (a) JPEG compression, (b) Gaussian noise, and (c) Gaussian blur.}}
    \label{fig:robustness}
    \vspace{-0.5cm}
\end{figure}

\paragraph{Robustness to Image Perturbations.}
In~\Cref{fig:robustness} we evaluate robustness under JPEG compression, Gaussian noise and Gaussian blur.
While accuracy remains stable for JPEG factors up to 30 (a), it shows a steady decay under Gaussian noise (b). Notably, Gaussian blur (c) induces a sharp initial drop at $\sigma=1$ followed by a plateau, suggesting that the model successfully pivots to structural cues when local textures are suppressed.





\paragraph{Out-of-Distribution Robustness.}
To evaluate the generalization capability of our method, we test FRIDA on several OOD datasets. 
Our approach achieves a solid average accuracy of 73.5\%, demonstrating consistent detection across diverse generative architectures. 
Specifically, the model yields 69.4\% on Flux 1.0 Dev, 68.1\% on Flux 1.1 Pro, 72.5\% on GPT Image 1, and 73.0\% on SDv3.5. 
We observe a significant performance peak on Qwen (95.3\%), while even on the most challenging generator in our benchmark, Ideogram 3.0, the detector maintains an accuracy of 62.7\%.

\vspace{-2em}
\begin{table}[h]
\centering
\scriptsize
\setlength{\tabcolsep}{3pt}
\resizebox{0.7\columnwidth}{!}{
\begin{tabular}{lccccccccc}
\toprule
Backbone & Midj. & SDv1.4 & SDv1.5 & ADM & GLIDE & BigGAN & VQDM & Wukong & Avg \\
\midrule
\rowcolor{yellow!25}
SD (Ours) & 88.8 & 87.3 & 87.3 & 87.2 & 90.2 & 87.2 & 88.8 & 87.2 & 88.0 \\
CLIP      & 63.6 & 64.6 & 64.5 & 57.9 & 66.3 & 60.2 & 55.9 & 64.1 & 61.5 \\
DINO      & 58.0 & 62.5 & 62.3 & 57.5 & 66.2 & 63.0 & 61.9 & 63.7 & 61.2 \\
\bottomrule
\end{tabular}}
\normalsize
\caption{\textbf{Backbone comparison.} $k$-NN fake image detection replacing SDM with ViT-based backbones (CLIP and DINO).}
\label{tab:backbone_comparison}
\vspace{-1.4cm}
\end{table}



%% file: sec/conclusions.tex
\section{Conclusions}
\label{sec:conclusions}
In this work, we introduced FRIDA, a framework that leverages the internal representation of a pre-trained diffusion model for fake image detection and source model attribution. 
For fake image detection, FRIDA entirely operates within the diffusion feature space, exploiting latent representations from the Stable Diffusion U-Net as compact and discriminative descriptors of image authenticity.
Through extensive evaluation, we demonstrated that a simple $k$-nearest neighbour classifier applied to these features achieves state-of-the-art detection performance with a limited support set, generalizing effectively to unseen generators without any retraining. 
For source model attribution, a compact neural classifier trained on the same features accurately identifies the generative source, confirming that diffusion features encode model-specific fingerprints. 
Furthermore, we show that diffusion-based features are more discriminative than CLIP and DINO features for fake image detection. 
Through extensive testing, we also demonstrate that our method maintains high performance under various image distortions and effectively generalizes to out-of-distribution generators, including state-of-the-art models like Flux and Qwen.
These findings may position diffusion representations as a reliable foundation for AI-generated image forensics. 

%% file: main.bib
@String(CVPR  = {IEEE Conf. Comput. Vis. Pattern Recog.})

@String(ECCV  = {Eur. Conf. Comput. Vis.})

@String(NeurIPS = {Adv. Neural Inform. Process. Syst.})

@String(ICASSP=	{ICASSP})

@String(CVPR  = {CVPR})

@String(ECCV  = {ECCV})

@String(NeurIPS = {NeurIPS})

@article{wu2025explainable,
  title={Explainable Synthetic Image Detection through Diffusion Timestep Ensembling},
  author={Wu, Yixin and Zhang, Feiran and Shi, Tianyuan and Yin, Ruicheng and Wang, Zhenghua and Gan, Zhenliang and Wang, Xiaohua and Lv, Changze and Zheng, Xiaoqing and Huang, Xuanjing},
  journal={arXiv preprint arXiv:2503.06201},
  year={2025}
}

@inproceedings{radford2021learning,
  title={Learning transferable visual models from natural language supervision},
  author={Radford, Alec and Kim, Jong Wook and Hallacy, Chris and Ramesh, Aditya and Goh, Gabriel and Agarwal, Sandhini and Sastry, Girish and Askell, Amanda and Mishkin, Pamela and Clark, Jack and others},
  booktitle={International conference on machine learning},
  pages={8748--8763},
  year={2021},
  organization={PmLR}
}

@article{zhu2023genimage,
  title={Genimage: A million-scale benchmark for detecting ai-generated image},
  author={Zhu, Mingjian and Chen, Hanting and Yan, Qiangyu and Huang, Xudong and Lin, Guanyu and Li, Wei and Tu, Zhijun and Hu, Hailin and Hu, Jie and Wang, Yunhe},
  journal={Advances in Neural Information Processing Systems},
  volume={36},
  pages={77771--77782},
  year={2023}
}

@misc{midjourney,
  title = {Midjourney},
  year = 2022,
  howpublished = {\url{https://www.midjourney.com/home/}}
}

@misc{nanobanana,
  title = {Gemini 2.5 Flash Image (codename “Nano Banana”)},
  author={Google DeepMind},
  year = 2025,
  howpublished = {\url{https://aistudio.google.com/}}
}

@article{betker2023improving,
  title={Improving image generation with better captions},
  author={Betker, James and Goh, Gabriel and Jing, Li and Brooks, Tim and Wang, Jianfeng and Li, Linjie and Ouyang, Long and Zhuang, Juntang and Lee, Joyce and Guo, Yufei and others},
  journal={Computer Science. https://cdn. openai. com/papers/dall-e-3. pdf},
  volume={2},
  number={3},
  pages={8},
  year={2023}
}

@article{KAZEROUNI2023102846,
title = {Diffusion models in medical imaging: A comprehensive survey},
journal = {Medical Image Analysis},
volume = {88},
pages = {102846},
year = {2023},
issn = {1361-8415},
doi = {https://doi.org/10.1016/j.media.2023.102846},
url = {https://www.sciencedirect.com/science/article/pii/S1361841523001068},
author = {Amirhossein Kazerouni and Ehsan Khodapanah Aghdam and Moein Heidari and Reza Azad and Mohsen Fayyaz and Ilker Hacihaliloglu and Dorit Merhof}
}

@article{WANG2025102593,
title = {Diffusion model for medical image denoising, reconstruction and translation},
journal = {Computerized Medical Imaging and Graphics},
volume = {124},
pages = {102593},
year = {2025},
issn = {0895-6111},
doi = {https://doi.org/10.1016/j.compmedimag.2025.102593},
url = {https://www.sciencedirect.com/science/article/pii/S0895611125001028},
author = {Wei Wang and Jiayu Xia and Gongning Luo and Suyu Dong and Xiangyu Li and Jie Wen and Shuo Li}
}

@article{10937272,
  author={Tivnan, Matthew and Teneggi, Jacopo and Lee, Tzu-Cheng and Zhang, Ruoqiao and Boedeker, Kirsten and Cai, Liang and Gang, Grace J. and Sulam, Jeremias and Stayman, J. Webster},
  journal={IEEE Transactions on Medical Imaging}, 
  title={Fourier Diffusion Models: A Method to Control MTF and NPS in Score-Based Stochastic Image Generation}, 
  year={2025},
  volume={44},
  number={9},
  pages={3694-3704},
  doi={10.1109/TMI.2025.3553805}}

@article{xia2024scoping,
  title={A scoping review on how generative artificial intelligence transforms assessment in higher education},
  author={Xia, Qi and Weng, Xiaojing and Ouyang, Fan and Lin, Tzung Jin and Chiu, Thomas KF},
  journal={International Journal of Educational Technology in Higher Education},
  volume={21},
  number={1},
  pages={40},
  year={2024},
  publisher={Springer}
}

@article{wu2025systematic,
  title={A Systematic Review of Responses, Attitudes, and Utilization Behaviors on Generative AI for Teaching and Learning in Higher Education},
  author={Wu, Fan and Dang, Yang and Li, Manli},
  journal={Behavioral Sciences},
  volume={15},
  number={4},
  pages={467},
  year={2025}
}

@article{grewal2025generative,
  title={How generative AI Is shaping the future of marketing},
  author={Grewal, Dhruv and Satornino, Cinthia B and Davenport, Thomas and Guha, Abhijit},
  journal={Journal of the Academy of Marketing Science},
  volume={53},
  number={3},
  pages={702--722},
  year={2025},
  publisher={Springer}
}

@article{cillo2025generative,
  title={Generative AI in innovation and marketing processes: A roadmap of research opportunities},
  author={Cillo, Paola and Rubera, Gaia},
  journal={Journal of the Academy of Marketing Science},
  volume={53},
  number={3},
  pages={684--701},
  year={2025},
  publisher={Springer}
}

@article{wolf2025diffusion,
  title={Diffusion models for robotic manipulation: A survey},
  author={Wolf, Rosa Petra and Shi, Yitian and Liu, Sheng and Rayyes, Rania},
  journal={Frontiers in Robotics and AI},
  volume={12},
  pages={1606247},
  year={2025},
  publisher={Frontiers}
}

@inproceedings{sha2023fake,
  title={De-fake: Detection and attribution of fake images generated by text-to-image generation models},
  author={Sha, Zeyang and Li, Zheng and Yu, Ning and Zhang, Yang},
  booktitle={Proceedings of the 2023 ACM SIGSAC conference on computer and communications security},
  pages={3418--3432},
  year={2023}
}

@inproceedings{luo2024lare,
  title={LaRE\^{} 2: Latent reconstruction error based method for diffusion-generated image detection},
  author={Luo, Yunpeng and Du, Junlong and Yan, Ke and Ding, Shouhong},
  booktitle={Proceedings of the IEEE/CVF Conference on Computer Vision and Pattern Recognition},
  pages={17006--17015},
  year={2024}
}

@inproceedings{wang2023dire,
  title={Dire for diffusion-generated image detection},
  author={Wang, Zhendong and Bao, Jianmin and Zhou, Wengang and Wang, Weilun and Hu, Hezhen and Chen, Hong and Li, Houqiang},
  booktitle={Proceedings of the IEEE/CVF International Conference on Computer Vision},
  pages={22445--22455},
  year={2023}
}

@article{BONECHI2025128846,
title = {An analysis of pre-trained stable diffusion models through a semantic lens},
journal = {Neurocomputing},
volume = {614},
pages = {128846},
year = {2025},
issn = {0925-2312},
doi = {https://doi.org/10.1016/j.neucom.2024.128846},
url = {https://www.sciencedirect.com/science/article/pii/S0925231224016175},
author = {Simone Bonechi and Paolo Andreini and Barbara Toniella Corradini and Franco Scarselli}
}

@article{BONECHI20245244,
title = {Diff-Props: is Semantics Preserved within a Diffusion Model?},
journal = {Procedia Computer Science},
volume = {246},
pages = {5244-5253},
year = {2024},
note = {28th International Conference on Knowledge Based and Intelligent information and Engineering Systems (KES 2024)},
issn = {1877-0509},
doi = {https://doi.org/10.1016/j.procs.2024.09.628},
url = {https://www.sciencedirect.com/science/article/pii/S1877050924026826},
author = {Simone Bonechi and Paolo Andreini and Barbara Toniella Corradini and Franco Scarselli}
}

@inproceedings{yu2019attributing,
  title={Attributing fake images to gans: Learning and analyzing gan fingerprints},
  author={Yu, Ning and Davis, Larry S and Fritz, Mario},
  booktitle={Proceedings of the IEEE/CVF international conference on computer vision},
  pages={7556--7566},
  year={2019}
}

@article{baranchuk2021label,
  title={Label-efficient semantic segmentation with diffusion models},
  author={Baranchuk, Dmitry and Rubachev, Ivan and Voynov, Andrey and Khrulkov, Valentin and Babenko, Artem},
  journal={arXiv preprint arXiv:2112.03126},
  year={2021}
}

@article{vasilcoiu2025latte,
  title={LATTE: Latent Trajectory Embedding for Diffusion-Generated Image Detection},
  author={Vasilcoiu, Ana and Najdenkoska, Ivona and Geradts, Zeno and Worring, Marcel},
  journal={arXiv preprint arXiv:2507.03054v1},
  year={2025}
}

@inproceedings{cazenavette2024fakeinversion,
  title={FakeInversion: Learning to Detect Images from Unseen Text-to-Image Models by Inverting Stable Diffusion},
  author={Cazenavette, George and Sud, Avneesh and Leung, Thomas and Usman, Ben},
  booktitle={Proceedings of the IEEE/CVF Conference on Computer Vision and Pattern Recognition},
  pages={10759--10769},
  year={2024}
}

@InProceedings{Rombach_2022_CVPR,
    author    = {Rombach, Robin and Blattmann, Andreas and Lorenz, Dominik and Esser, Patrick and Ommer, Bj\"orn},
    title     = {High-Resolution Image Synthesis With Latent Diffusion Models},
    booktitle = {Proceedings of the IEEE/CVF Conference on Computer Vision and Pattern Recognition (CVPR)},
    month     = {June},
    year      = {2022},
    pages     = {10684-10695}
}

@article{lundberg2017unified,
  title={A unified approach to interpreting model predictions},
  author={Lundberg, Scott M and Lee, Su-In},
  journal={Advances in neural information processing systems},
  volume={30},
  year={2017}
}

@article{brock2018large,
  title={Large Scale GAN Training for High Fidelity Natural Image Synthesis},
  author={Brock, Andrew},
  journal={arXiv preprint arXiv:1809.11096},
  year={2018}
}

@inproceedings{nichol2022glide,
  title={GLIDE: Towards Photorealistic Image Generation and Editing with Text-Guided Diffusion Models},
  author={Nichol, Alexander Quinn and Dhariwal, Prafulla and Ramesh, Aditya and Shyam, Pranav and Mishkin, Pamela and Mcgrew, Bob and Sutskever, Ilya and Chen, Mark},
  booktitle={International Conference on Machine Learning},
  pages={16784--16804},
  year={2022},
  organization={PMLR}
}

@inproceedings{gu2022vector,
  title={Vector quantized diffusion model for text-to-image synthesis},
  author={Gu, Shuyang and Chen, Dong and Bao, Jianmin and Wen, Fang and Zhang, Bo and Chen, Dongdong and Yuan, Lu and Guo, Baining},
  booktitle={Proceedings of the IEEE/CVF conference on computer vision and pattern recognition},
  pages={10696--10706},
  year={2022}
}

@misc{wukong,
  title = {Wukong},
  year = 2022,
  howpublished = {\url{https://xihe.mindspore.cn/modelzoo/wukong/introduce}}
}

@inproceedings{deng2009imagenet,
  title={Imagenet: A large-scale hierarchical image database},
  author={Deng, Jia and Dong, Wei and Socher, Richard and Li, Li-Jia and Li, Kai and Fei-Fei, Li},
  booktitle={2009 IEEE conference on computer vision and pattern recognition},
  pages={248--255},
  year={2009},
  organization={Ieee}
}

@article{dhariwal2021diffusion,
  title={Diffusion models beat gans on image synthesis},
  author={Dhariwal, Prafulla and Nichol, Alexander},
  journal={Advances in neural information processing systems},
  volume={34},
  pages={8780--8794},
  year={2021}
}

@article{corradini2024freeseg,
  title={Freeseg-diff: Training-free open-vocabulary segmentation with diffusion models},
  author={Corradini, Barbara Toniella and Shukor, Mustafa and Couairon, Paul and Couairon, Guillaume and Scarselli, Franco and Cord, Matthieu},
  journal={arXiv preprint arXiv:2403.20105},
  year={2024}
}

@InProceedings{diffimplicit2025,
author="Wang, Xi
and Kalogeiton, Vicky",
editor="Del Bue, Alessio
and Canton, Cristian
and Pont-Tuset, Jordi
and Tommasi, Tatiana",
title="Your Diffusion Model is an Implicit Synthetic Image Detector",
booktitle="Computer Vision -- ECCV 2024 Workshops",
year="2025",
publisher="Springer Nature Switzerland",
address="Cham",
pages="418--434",
isbn="978-3-031-92648-8"
}

@InProceedings{clipfeat2024,
author="Cioni, Dario
and Tzelepis, Christos
and Seidenari, Lorenzo
and Patras, Ioannis",
editor="Del Bue, Alessio
and Canton, Cristian
and Pont-Tuset, Jordi
and Tommasi, Tatiana",
title="Are CLIP Features All You Need for Universal Synthetic Image Origin Attribution?",
booktitle="Computer Vision -- ECCV 2024 Workshops",
year="2025",
publisher="Springer Nature Switzerland",
address="Cham",
pages="363--382",
isbn="978-3-031-92648-8"
}

@inproceedings{defake2023,
author = {Sha, Zeyang and Li, Zheng and Yu, Ning and Zhang, Yang},
title = {DE-FAKE: Detection and Attribution of Fake Images Generated by Text-to-Image Generation Models},
year = {2023},
isbn = {9798400700507},
publisher = {Association for Computing Machinery},
address = {New York, NY, USA},
url = {https://doi.org/10.1145/3576915.3616588},
doi = {10.1145/3576915.3616588},
booktitle = {Proceedings of the 2023 ACM SIGSAC Conference on Computer and Communications Security},
pages = {3418–3432},
numpages = {15},
location = {Copenhagen, Denmark},
series = {CCS '23}
}

@inproceedings{bui2022repmix,
  title={Repmix: Representation mixing for robust attribution of synthesized images},
  author={Bui, Tu and Yu, Ning and Collomosse, John},
  booktitle={European Conference on Computer Vision},
  pages={146--163},
  year={2022},
  organization={Springer}
}

@inproceedings{wang2020cnn,
  title={CNN-generated images are surprisingly easy to spot... for now},
  author={Wang, Sheng-Yu and Wang, Oliver and Zhang, Richard and Owens, Andrew and Efros, Alexei A},
  booktitle={Proceedings of the IEEE/CVF conference on computer vision and pattern recognition},
  pages={8695--8704},
  year={2020}
}

@inproceedings{zhu2025maid,
  title={MAID: Model Attribution via Inverse Diffusion},
  author={Zhu, Luyu and Ye, Kai and Yao, Jiayu and Li, Chenxi and Zhao, Luwen and Cao, Yuxin and Wang, Derui and Hao, Jie},
  booktitle={ICASSP 2025-2025 IEEE International Conference on Acoustics, Speech and Signal Processing (ICASSP)},
  pages={1--5},
  year={2025},
  organization={IEEE}
}

@inproceedings{semgir2024,
author = {Yu, Xiao and Chen, Kejiang and Zeng, Kai and Fang, Han and Yang, Zijin and Shang, Xiuwei and Qi, Yuang and Zhang, Weiming and Yu, Nenghai},
title = {SemGIR: Semantic-Guided Image Regeneration Based Method for AI-generated Image Detection and Attribution},
year = {2024},
isbn = {9798400706868},
publisher = {Association for Computing Machinery},
address = {New York, NY, USA},
url = {https://doi.org/10.1145/3664647.3680776},
doi = {10.1145/3664647.3680776},
booktitle = {Proceedings of the 32nd ACM International Conference on Multimedia},
pages = {8480–8488},
numpages = {9},
location = {Melbourne VIC, Australia},
series = {MM '24}
}

@inproceedings{nguyen2024exploring,
  title={Exploring self-supervised vision transformers for deepfake detection: A comparative analysis},
  author={Nguyen, Huy H and Yamagishi, Junichi and Echizen, Isao},
  booktitle={2024 IEEE International Joint Conference on Biometrics (IJCB)},
  pages={1--10},
  year={2024},
  organization={IEEE}
}

@inproceedings{smeu2025declip,
  title={DeCLIP: Decoding CLIP representations for deepfake localization},
  author={Smeu, Stefan and Oneata, Elisabeta and Oneata, Dan},
  booktitle={2025 IEEE/CVF Winter Conference on Applications of Computer Vision (WACV)},
  pages={149--159},
  year={2025},
  organization={IEEE}
}

@inproceedings{wang2025fsfm,
  title={Fsfm: A generalizable face security foundation model via self-supervised facial representation learning},
  author={Wang, Gaojian and Lin, Feng and Wu, Tong and Liu, Zhenguang and Ba, Zhongjie and Ren, Kui},
  booktitle={Proceedings of the Computer Vision and Pattern Recognition Conference},
  pages={24364--24376},
  year={2025}
}

@inproceedings{caron2021emerging,
  title={Emerging properties in self-supervised vision transformers},
  author={Caron, Mathilde and Touvron, Hugo and Misra, Ishan and J{\'e}gou, Herv{\'e} and Mairal, Julien and Bojanowski, Piotr and Joulin, Armand},
  booktitle={Proceedings of the IEEE/CVF international conference on computer vision},
  pages={9650--9660},
  year={2021}
}

@inproceedings{liu2020global,
  title={Global texture enhancement for fake face detection in the wild},
  author={Liu, Zhengzhe and Qi, Xiaojuan and Torr, Philip HS},
  booktitle={Proceedings of the IEEE/CVF conference on computer vision and pattern recognition},
  pages={8060--8069},
  year={2020}
}

@inproceedings{zhang2019detecting,
  title={Detecting and simulating artifacts in gan fake images},
  author={Zhang, Xu and Karaman, Svebor and Chang, Shih-Fu},
  booktitle={2019 IEEE international workshop on information forensics and security (WIFS)},
  pages={1--6},
  year={2019},
  organization={IEEE}
}

@inproceedings{qian2020thinking,
  title={Thinking in frequency: Face forgery detection by mining frequency-aware clues},
  author={Qian, Yuyang and Yin, Guojun and Sheng, Lu and Chen, Zixuan and Shao, Jing},
  booktitle={European conference on computer vision},
  pages={86--103},
  year={2020},
  organization={Springer}
}

@article{saharia2022photorealistic,
  title={Photorealistic text-to-image diffusion models with deep language understanding},
  author={Saharia, Chitwan and Chan, William and Saxena, Saurabh and Li, Lala and Whang, Jay and Denton, Emily L and Ghasemipour, Kamyar and Gontijo Lopes, Raphael and Karagol Ayan, Burcu and Salimans, Tim and others},
  journal={Advances in neural information processing systems},
  volume={35},
  pages={36479--36494},
  year={2022}
}

@inproceedings{kerssies2025your,
  title={Your vit is secretly an image segmentation model},
  author={Kerssies, Tommie and Cavagnero, Niccolo and Hermans, Alexander and Norouzi, Narges and Averta, Giuseppe and Leibe, Bastian and Dubbelman, Gijs and de Geus, Daan},
  booktitle={Proceedings of the Computer Vision and Pattern Recognition Conference},
  pages={25303--25313},
  year={2025}
}

@inproceedings{
dosovitskiy2021an,
title={An Image is Worth 16x16 Words: Transformers for Image Recognition at Scale},
author={Alexey Dosovitskiy and Lucas Beyer and Alexander Kolesnikov and Dirk Weissenborn and Xiaohua Zhai and Thomas Unterthiner and Mostafa Dehghani and Matthias Minderer and Georg Heigold and Sylvain Gelly and Jakob Uszkoreit and Neil Houlsby},
booktitle={International Conference on Learning Representations},
year={2021},
url={https://openreview.net/forum?id=YicbFdNTTy}
}

@inproceedings{wang2024enhancing,
  title={Enhancing pre-trained vits for downstream task adaptation: A locality-aware prompt learning method},
  author={Wang, Shaokun and Yu, Yifan and He, Yuhang and Gong, Yihong},
  booktitle={Proceedings of the 32nd ACM International Conference on Multimedia},
  pages={797--806},
  year={2024}
}

@inproceedings{zhong2025beyond,
  title={Beyond Generation: A Diffusion-based Low-level Feature Extractor for Detecting AI-generated Images},
  author={Zhong, Nan and Chen, Haoyu and Xu, Yiran and Qian, Zhenxing and Zhang, Xinpeng},
  booktitle={Proceedings of the Computer Vision and Pattern Recognition Conference},
  pages={8258--8268},
  year={2025}
}

@inproceedings{
loshchilov2018decoupled,
title={Decoupled Weight Decay Regularization},
author={Ilya Loshchilov and Frank Hutter},
booktitle={International Conference on Learning Representations},
year={2019},
url={https://openreview.net/forum?id=Bkg6RiCqY7},
}

@inproceedings{ramesh2021zero,
  title={Zero-shot text-to-image generation},
  author={Ramesh, Aditya and Pavlov, Mikhail and Goh, Gabriel and Gray, Scott and Voss, Chelsea and Radford, Alec and Chen, Mark and Sutskever, Ilya},
  booktitle={International conference on machine learning},
  pages={8821--8831},
  year={2021},
  organization={Pmlr}
}

@article{szekely2007distance,
  title = {Distance correlation: a measure for dependence between multivariate random variables},
  author = {Székely, Gábor J and Rizzo, Maria L},
  journal = {Annals of Statistics},
  volume = 35,
  number = 6,
  pages = {2769--2792},
  year = {2007},
  publisher = {Institute of Mathematical Statistics}
}

@article{krause1973taxicab,
  title={Taxicab geometry},
  author={Krause, Eugene F},
  journal={The Mathematics Teacher},
  volume={66},
  number={8},
  pages={695--706},
  year={1973},
  publisher={JSTOR}
}

@book{manning2008introduction,
  title={Introduction to information retrieval},
  author={Manning, Christopher D. and Raghavan, Prabhakar and Sch{\"u}tze, Hinrich},
  year={2008},
  publisher={Cambridge university press}
}

@inproceedings{chen2020simple,
  title={A simple framework for contrastive learning of visual representations},
  author={Chen, Ting and Kornblith, Simon and Norouzi, Mohammad and Hinton, Geoffrey},
  booktitle={International conference on machine learning},
  pages={1597--1607},
  year={2020},
  organization={PmLR}
}

@inproceedings{he2020momentum,
  title={Momentum contrast for unsupervised visual representation learning},
  author={He, Kaiming and Fan, Haoqi and Wu, Yuxin and Xie, Saining and Girshick, Ross},
  booktitle={Proceedings of the IEEE/CVF conference on computer vision and pattern recognition},
  pages={9729--9738},
  year={2020}
}

@inproceedings{cioni2024clip,
  title={Are CLIP features all you need for Universal Synthetic Image Origin Attribution?},
  author={Cioni, Dario and Tzelepis, Christos and Seidenari, Lorenzo and Patras, Ioannis},
  booktitle={European Conference on Computer Vision},
  pages={363--382},
  year={2024},
  organization={Springer}
}

@misc{wu2025qwenimagetechnicalreport,
      title={Qwen-Image Technical Report}, 
      author={Chenfei Wu and Jiahao Li and Jingren Zhou and Junyang Lin and Kaiyuan Gao and Kun Yan and Sheng-ming Yin and Shuai Bai and Xiao Xu and Yilei Chen and Yuxiang Chen and Zecheng Tang and Zekai Zhang and Zhengyi Wang and An Yang and Bowen Yu and Chen Cheng and Dayiheng Liu and Deqing Li and Hang Zhang and Hao Meng and Hu Wei and Jingyuan Ni and Kai Chen and Kuan Cao and Liang Peng and Lin Qu and Minggang Wu and Peng Wang and Shuting Yu and Tingkun Wen and Wensen Feng and Xiaoxiao Xu and Yi Wang and Yichang Zhang and Yongqiang Zhu and Yujia Wu and Yuxuan Cai and Zenan Liu},
      year={2025},
      eprint={2508.02324},
      archivePrefix={arXiv},
      primaryClass={cs.CV},
      url={https://arxiv.org/abs/2508.02324}, 
}

@article{livernoche2025openfake,
  title={OpenFake: An Open Dataset and Platform Toward Real-World Deepfake Detection},
  author={Livernoche, Victor and Arodi, Akshatha and Musulan, Andreea and Yang, Zachary and Salvail, Adam and Caron, Ga{\'e}tan Marceau and Godbout, Jean-Fran{\c{c}}ois and Rabbany, Reihaneh},
  journal={arXiv preprint arXiv:2509.09495},
  year={2025}
}

@inproceedings{schuhmann2021laion,
  title={LAION-400M: Open Dataset of CLIP-Filtered 400 Million Image-Text Pairs},
  author={Schuhmann, Christoph and Kaczmarczyk, Robert and Komatsuzaki, Aran and Katta, Aarush and Vencu, Richard and Beaumont, Romain and Jitsev, Jenia and Coombes, Theo and Mullis, Clayton},
  booktitle={NeurIPS Workshop Datacentric AI},
  number={FZJ-2022-00923},
  year={2021},
  organization={J{\"u}lich Supercomputing Center}
}

@Article{Everingham15,
   author = "Everingham, M. and Eslami, S. M. A. and Van~Gool, L. and Williams, C. K. I. and Winn, J. and Zisserman, A.",
   title = "The Pascal Visual Object Classes Challenge: A Retrospective",
   journal = "International Journal of Computer Vision",
   volume = "111",
   year = "2015",
   number = "1",
   month = jan,
   pages = "98--136",
}
